%% file: main.tex
\documentclass[journal]{IEEEtran} 

\IEEEoverridecommandlockouts

\usepackage{amssymb, amsmath, amsfonts} 
\usepackage{mathtools}      
\usepackage{enumitem}       
\usepackage{graphicx}       
\usepackage{latexsym}
\usepackage{xcolor}         
\usepackage{blindtext}      
\usepackage{stfloats}       
\usepackage{subfigure}
\definecolor{myblue}{rgb}{0.0, 0.0, 0.60}
\usepackage[colorlinks=true, linkcolor=myblue, citecolor=myblue, urlcolor=myblue]{hyperref}

\usepackage{blindtext}
\usepackage{stfloats}
\usepackage{mathtools}
\usepackage{svg}
\usepackage{amsmath,amssymb,amsfonts}
\usepackage{algorithmic}
\usepackage{graphicx}
\usepackage{textcomp}
\usepackage{xcolor}
\usepackage[nameinlink]{cleveref}   
\crefname{figure}{Fig.}{figures}
\usepackage[
    backend=biber,
    style=ieee,
    sorting=none
]{biblatex}
\begin{document}
\title{Learning to Move Objects with Fluid Streams \\ in a Differentiable Simulation


\thanks{This research is funded by the Latvian Council of Science, project “Smart Materials, Photonics, Technologies and Engineering Ecosystem”, project No. VPP-EM-FOTONIKA-2022/1-0001.}
}
\author{
    \IEEEauthorblockN{Kārlis Freivalds},
    \IEEEauthorblockN{Laura Leja},
    \IEEEauthorblockN{Oskars Teikmanis}
    \\
    \IEEEauthorblockA{\textit{Institute of Electronics and Computer Science (EDI), Riga, Latvia}}\\
    \IEEEauthorblockA{\texttt{\{karlis.freivalds, laura.leja, oskars.teikmanis\}@edi.lv}}
}
\maketitle

\begin{abstract}
 \input{sections/0_abstract}

\end{abstract}


\input{sections/1_introduction}

\input{sections/2_method}

\input{sections/3_results}

\input{sections/4_conclusion}

\printbibliography

\end{document}

%% file: sections/0_abstract.tex
We introduce a method for manipulating objects in three-dimensional space using controlled fluid streams. To achieve this, we train a neural network controller in a differentiable simulation and evaluate it in a simulated environment consisting of an $8\times8$ grid of vertical emitters. By carrying out various horizontal displacement tasks such as moving objects to specific positions while reacting to external perturbations, we demonstrate that a controller, trained with a limited number of iterations, can generalise to longer episodes and learn the complex dynamics of fluid-solid interactions. Importantly, our approach requires only the observation of the manipulated object's state, paving the way for the development of physical systems that enable contactless manipulation of objects using air streams.

%% file: sections/1_introduction.tex
\section{Introduction}

Real-time control of systems involving fluid and solid interactions presents inherent challenges due to the complex dynamics arising from their interactions and the high-dimensional state space. Designing a controller for such a system, for example, one that moves an object to a prescribed location with a fluid jet, is highly impractical through manual programming alone. However, by leveraging a suitable simulator in conjunction with modern deep learning techniques, it becomes feasible to train such a controller, which is the focus of this paper. We present an approach for controlling fluids using differentiable physics (DP) that only requires monitoring the state of the displaced solids, making it highly applicable to the design of physical fluid-based control systems.

Advances in DP \cite{thuerey2021pbdl} have demonstrated significant potential for addressing simulated inverse problems and complex control applications \cite{hu2019difftaichi, holl2020learning, du2021_diffpd, fang2022complex, ramos2022control, teikmanis2023applying, tathawadekar2023incomplete}. In the context of computational fluid-solid interactions, there are several examples of applied DP \cite{ramos2022control, ma2021diffaqua, hu2019difftaichi, du2021_diffpd, li2023difffr}. However, to our knowledge, there is no existing research of DP in control tasks in which the fluid itself serves as the acting agent.

\begin{figure}[t!]
    \centering
    \includegraphics[width=0.8\linewidth]{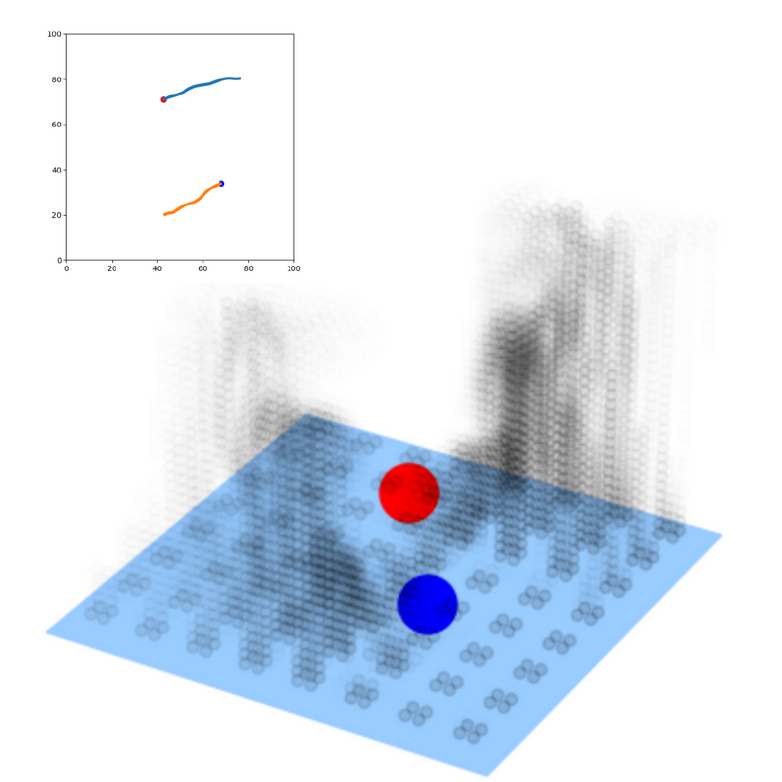}
    \caption{\textbf{Controlled displacement of simulated objects}. The solid lines in the top left shows the trajectory of the objects as they are moved by the simulated fluid. The fluid is emitted from an $8\times8$ grid, and is visualised as a grey plume. Animations can be found in the accompanied \href{https://www.youtube.com/watch?v=sft0MH_pk9w}{video}.}
    \label{fig:title}
\end{figure}

We propose a scenario in which controlled fluid streams are employed to displace objects in three-dimensional space with a grid of emitters. Using differentiable simulations, we train a neural network that produces a fluid velocity for each emitter such that objects are moved to a desired location, as shown in Fig. \ref{fig:title}. The system observes only the state (position, velocity) of the objects to carry out its control policy. Consequently, unlike prior approaches that rely on the fluid state (velocity and pressure fields) for decision-making \cite{ma2018fluid, ren2022versatile}, our approach has the potential for a real-world implementation.

\textbf{Our key contributions} are the following: (1) present a sample-efficient DP-based learning method to train a neural network controller for moving objects around a horizontal surface with fixed vertical emitters, using only the state of the object as a control input, and (2) demonstrate versatile horizontal object displacement using fixed vertical emitters in a simulated environment.

The remainder of this paper is structured as follows: Section~\ref{sec:method} describes the simulation environment and governing fluid dynamics equations. Section~\ref{sec:NN} offers insights into our controller design, and Section~\ref{sec:training} describes the training process. An evaluation of all experiments we performed with the controller is available in Section~\ref{sec:results}. Lastly, Section~\ref{sec:conclusion} puts the developed system into perspective, discussing its limitations and applicability in real-life counterparts.

\subsection{Related Work}

Through the application of gradient-based optimisation, DP has been effectively applied to solve simulated inverse problems \cite{hu2019difftaichi, holl2020learning, du2021_diffpd}. Hu et al. \cite{hu2019difftaichi} leverage the differentiable programming language Taichi to conduct complex simulations involving multi-contact interactions between rigid and soft bodies, as well as computational fluid dynamics. Du et al. \cite{du2021_diffpd} present a dedicated differentiable simulator tailored for soft-body manipulation and showcase its use in system identification, trajectory planning and real-to-simulation transfer. Holl et al. \cite{holl2020learning} introduce PhiFlow, a differentiable PDE solver that can be used to handle control and planning tasks for high-dimensional problems, including Navier-Stokes equations for incompressible fluids. Among potential solvers, PhiFlow offers a good fit for our application.

DP has also proven effective in diverse control applications \cite{fang2022complex, ramos2022control, teikmanis2023applying}. Ramos et al. \cite{ramos2022control} investigate a simulation involving a two-dimensional rigid object immersed in a fluid, and develop a DP-based controller for trajectory following amidst external perturbations. In a comparative analysis against a reference PID controller and a loop-shaping controller, the DP-based controller outperformed both counterparts, demonstrating lower average errors and reduced oscillations. The work of Ma et al. \cite{ma2021diffaqua} demonstrates a differentiable pipeline for designing the geometry and controller of simulated swimmers. By applying gradient-based parameter optimization, their proposed method results in significantly faster convergence and better loss than gradient-free alternatives.

Reinforcement learning (RL) has been effectively applied in designing controllers for fluid-solid interactions \cite{garnier2021review}, including drag reduction through active flow control \cite{rabault2019artificial, fan2020reinforcement}, efficient swarm-swimming strategies \cite{verma2018efficient} and fluid-directed rigid body control \cite{ma2018fluid, ren2022versatile}. In earlier work by Ma et al. \cite{ma2018fluid}, the objective is to balance and displace 2D objects with simulated air flows. In this example, the controller uses the state of the object, the emitter and the fluid itself as inputs, and the loss is a function of the object's deviation from the desired state. Although this approach proved successful in a simulated environment, the real-time measurement of the fluid state becomes highly impractical if it were to be applied on a physical device. Ren et al. \cite{ren2022versatile} offer to improve upon Ma et al. through the application of an off-policy RL algorithm (Soft Actor-Critic). While this change does improve the sample-efficiency of the method for completing similar tasks, the overall approach still requires the state of the fluid as an input to the controller. 

RL methods usually require a large number of samples, which are computationally expensive to obtain in 3D fluid simulations. In addition, they require careful parameter tuning to avoid unstable behaviour \cite{ren2022versatile}, and are challenging to scale to high dimensional parameter space. For instance, for their rigid body balancing scenario, Ma et al. \cite{ma2018fluid} train a system with an action space of 3 (spout position, angle and the on/off state of the jet), while our setup requires 64 continuous parameters that control the emitted fluid velocities.

In this paper, we aim to demonstrate the effectiveness of DP-based learning of three-dimensional fluid-directed rigid body control by applying gradient-based optimisation for simulated fluid-solid interactions, and showcase the resulting controller for enabling complex manipulation tasks by using only vertical fluid emitters, \textit{without} requiring any inputs relating to the state of the fluid during control.

%% file: sections/2_method.tex
\section{Simulation} \label{sec:method}

In this work we consider a scenario where objects are displaced with vertical fluid emitters arranged in a grid on a horizontal grid (also referred to as "table"), as shown in Fig.~\ref{fig:title}. The fluid velocities are controlled by a neural network controller, which takes the position and velocity of the objects as inputs, and outputs the required velocities for each emitter to guide the objects to their target locations.

The simulation domain is designed as a cuboid of size $100\times100\times100$ units with all boundaries open, except for the bottom boundary, which functions as a solid surface. On this surface, emitters are arranged in an $8\times8$ grid, each capable of emitting fluid at different velocities. Within this setup, one or more rigid objects are placed to demonstrate controlled solid-fluid interactions. Section~\ref{sec:results} features examples with one and two spherical objects with a radius of 10 units, but the approach can handle any number of such objects. 
 
The system behaviour is modelled using fluid dynamics within domain $\Omega$, governed by the well-known incompressible Navier-Stokes (NS) equations, applicable for time and space instances $x,t \in \Omega_f \times [0, t]$: 

\begin{equation} \label{eq:ns1}
\frac{\partial u}{\partial t} = -u \nabla u - \frac{\nabla p}{\rho} + \nu \nabla^2 u + g, \\
\end{equation}
\begin{equation} \label{eq:ns2}
\nabla \cdot u = 0,
\end{equation}
where the fluid velocity $u(t,x)$ and pressure $p(t,x)$ are spatiotemporal functions defined within $\Omega_f \subset \mathbb{ R}^2$. The term $\frac{\partial u}{\partial t}$ indicates the rate of change of fluid velocity over time. The term $-u \cdot \nabla u$ is the convective acceleration of the fluid. The pressure gradient force $-\frac{\nabla p}{\rho}$, arising from changes in pressure within the flow, is normalised by the fluid density $\rho$. The term $\nu \nabla^2 u$ represents the viscous diffusion of the velocity, where $\nu$ is the kinematic viscosity of the fluid, and $\nabla^2 u$ denotes the diffusion of the velocity field due to viscosity. The term $g$ represents external forces per unit mass acting on the fluid, such as gravity in our context.

The divergence constraint of the velocity field, $\nabla \cdot u = 0$, ensures the conservation of fluid mass and constant density within the flow. The negative sign in the pressure gradient term, $-\frac{\nabla p}{\rho}$, reflects a pressure drop in the direction of fluid flow, which is proportional to a given volumetric force $f$.

For our task of purposefully moving objects with fluid jets, it is essential to effectively manage fluid-solid interactions. This necessitates to account for the velocity and acceleration of the solids created by the fluid. To address this, we adapt the method from \cite{ramos2022control} which can be described with the following equations:

\begin{align}  
m \frac{\partial \mathbf{v}}{\partial t} 
  &= \mathbf{f} = - \oint_{\partial \Omega} p(x,t) \vec{n}(s) \, ds \label{eq:ns3} \\
  I \frac{\partial \mathbf{\alpha}}{\partial t} 
  &= \mathbf{s} = - \oint_{\partial \Omega} p(x,t) \vec{r}(s) \times \vec{n}(s) \, ds \label{eq:ns4}
  \end{align}
where \( m \) is the body mass, \( \mathbf{v} \) is the velocity of the body, \( \frac{\partial \mathbf{v}}{\partial t} \) is the acceleration, \( I \) is the moment of inertia, \( \mathbf{\omega} \) is the angular velocity, and \( \frac{\partial \mathbf{\omega}}{\partial t} \) represents the angular acceleration. The forces (\( \mathbf{f} \)) and torques (\( \mathbf{s} \)) acting on the body are calculated by integrating the pressure \( p(x,t) \) over the surface of the body \( \partial \Omega \), with \( \vec{n}(s) \) being the outward normal at the surface element \( ds \), and \( \vec{r}(s) \) representing the position vector relative to the centre of mass.

In parallel, the velocity of the fluid at the boundary of the region $\Omega$ equals the velocity of the immersed body, expressed as $\mathbf{v} + \mathbf{r} \times \boldsymbol{\alpha}$. In NS simulations, Dirichlet boundary conditions are imposed on the velocity field. These conditions are specified as $\mathbf{u}_\Omega = \frac{\partial \mathbf{v}}{\partial t} + \mathbf{r}_\Omega \frac{\partial \boldsymbol{\alpha}}{\partial t}$ for the rigid body surface. Our approach involves discretizing the fluid region at each time step, where the discretization must accurately describe the boundary surface of the body.

Our simulation environment is implemented using the differentiable PDE solver and simulation toolkit PhiFlow \cite{holl2020learning}. It supports fluid simulations and one-way coupling with solids, i.e., solid objects influence the fluid but not vice-versa. To enable full coupling, we employ the implementation from Ramos et al. \cite{ramos2022control}, adapted for 3D simulations. With this approach, we can calculate the force exerted by the fluid on solids. For the current implementation, we simulate only spherical objects, for which we implement Newtonian motion and collision handling using PhiFlow's integrated solver. 

The domain is discretized at a $32\times32\times32$ resolution, which is the maximum for training on one A100 GPU card. The time step is set to 0.5, representing the upper limit for maintaining training stability. The emitters are modelled as boundary conditions on the bottom plane of the domain. Each emitter functions as an inlet of a small area, producing fluid velocities as prescribed by the controller for each time step.

\begin{figure}[htbp!]
    \includegraphics[width=0.9\linewidth]{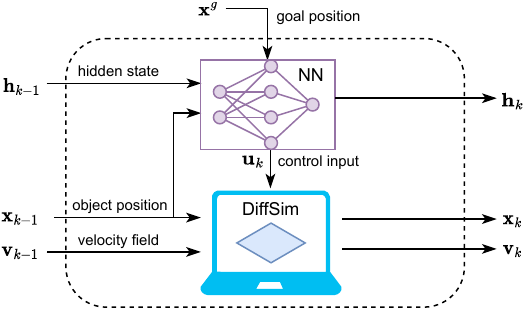}
    \caption{\textbf{Differentiable control diagram}. Shown is one time step of the controlled simulation. The differentiable simulator (bottom) operates on the velocity field $\mathbf{v}$ containing fluid velocity vectors for each spatial location at a time step $k$, the object's 3D position $\mathbf{x}$, and the control signal $\mathbf{u}$, corresponding to the emitter velocity. The neural network controller (top) receives the object's position from the simulator at the previous time step ($k-1$), its own hidden state $\mathbf{h}$, and the goal position of the controlled objects in order to update its hidden state and the control signal.}
    \label{fig:architecture}
\end{figure}

\section{Neural Network Controller Design}\label{sec:NN}

\begin{figure}[htbp!]
    \includegraphics[width=1\linewidth]{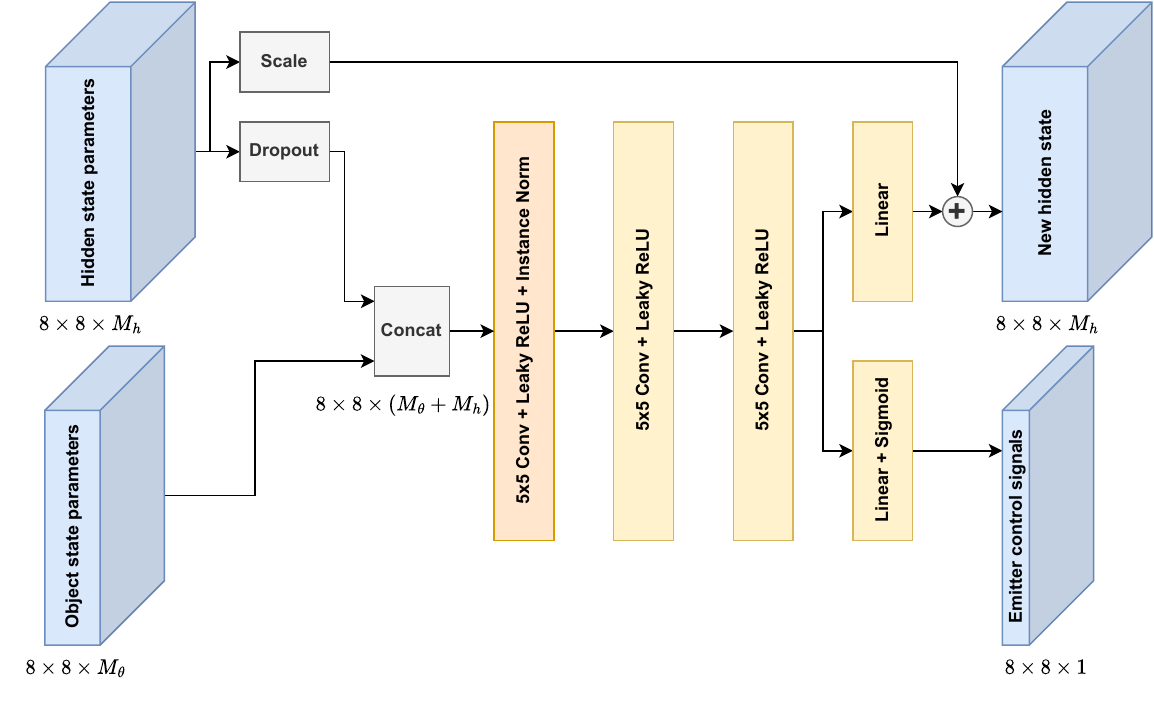}
    \caption{
    \textbf{Neural network architecture}. We use a recurrent convolutional neural network that operates on an input grid of $8\times8$ fluid emitters with a corresponding feature map of $M_{\theta}$ elements, and a hidden state of size $M_h$. Dropout is applied to the hidden state to enhance generalisation to longer simulations. Two output heads performing linear projection are attached to the last layer, producing emitter velocities (range-limited via sigmoid activation) and an updated hidden state for the next time step by applying a scaled residual connection.}
    \label{fig:NN}
\end{figure}

In a control scenario, the process depicted in Fig.~\ref{fig:architecture} is applied for each time step. The simulation is initialised with a velocity field $\mathbf{v}_0=\mathbf{0}$. For the hidden state vector $\mathbf{h}_0$, all elements are initialised with $1/\sqrt{M_h}$, a small nonzero value that prevents catastrophic errors during normalisation, which could occur if all values were zero inside the neural network depicted in Fig.~\ref{fig:NN} . The initial state $\mathbf{x}_0$ depends on the control task, but we usually set the initial velocity to zero and randomise the initial position within the boundaries of the environment. During training, this structure is unrolled through time, and the loss function is calculated depending on $\mathbf{x}_N$, where $N$ is the number of time steps. Gradient backpropagation is performed to adjust the neural network controller's parameters to minimise the loss.

The neural network architecture is depicted in Fig.~\ref{fig:NN}. It is designed as a recurrent convolutional network, receiving the object state representation and its own internal state as inputs at each time step. It produces an emitter control signal matrix and the next internal state tensor as outputs. It operates on spatial data with an $8\times8$ shape, corresponding to the emitter grid. By design, convolutions are well suited for such spatial data processing. Recurrence is implemented as scaled residual connections similar to ResNet \cite{he2016deep}, facilitating stable training through many time steps.

\begin{figure}[htbp!]
    \centering
    \hfill
    \subfigure[]{
        \includegraphics[width=0.45\linewidth]{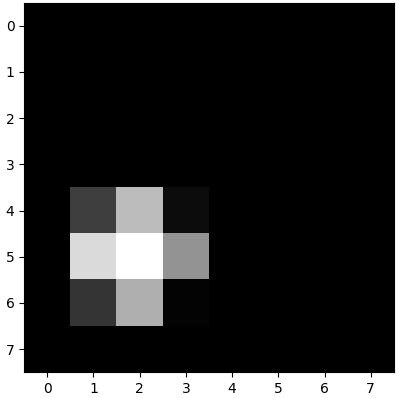} 
        \label{fig:object_map}
    }
    \hfill
    \subfigure[]{
        \includegraphics[width=0.45\linewidth]{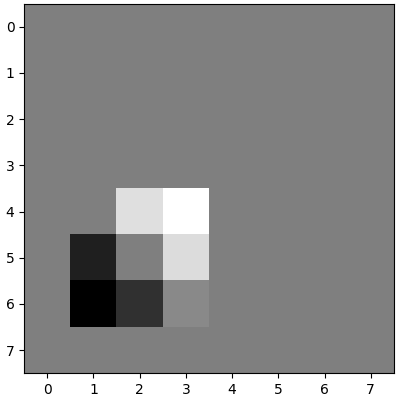} 
        \label{fig:velocity_map}
    }
    \hfill
    \caption{\textbf{Object position and velocity representation.} (a) Object soft mask rendered onto an $8\times8$ grid. A brighter shade indicates greater coverage of the object within a grid cell. (b) Velocity representation derived by subtracting two object masks displaced by velocity. Bright colours indicate positive values, dark colours represent negative values, and zero is depicted in grey.}
    \label{fig:object_maps}
\end{figure}

To fit the convolutional structure, the object state information is provided to the neural network as an $8\times8$ matrix with several feature maps: 
\begin{itemize}
    \item[1.]  The object's position in the $xy$ plane is represented as an $8\times8$ image onto which the object is rendered, see Fig.~\ref{fig:object_map}. To support gradient backpropagation, the object is rendered in PhiFlow using soft boundaries giving the volume fraction occupied by the object, similar to \cite{takahashi2021differentiable}. 
    \item[2.] The object’s $z$ coordinate is represented as the position mask scaled with the object's $z$ value.
    \item[3.] To represent the $x$ and $y$ components of the object's velocity, the difference of two object masks is calculated -- one from the current time step and another from the subsequent time step, as illustrated in Fig.~\ref{fig:velocity_map}. The position mask for the next time step is estimated by adding the current velocity to the current object position.
    \item[4.] For the $z$ component of the velocity, the object mask obtained in step 1 scaled by the velocity's $z$ component is used.
\end{itemize}
In addition, the goal position $\mathbf{x}^g$ is passed to the neural network as a distance field. Consequently, each spatial location $(i,j)$ within the convolutional map is assigned the absolute value of the distance to the goal, calculated as  $|x^g-(i,j,0)|$ for the $x$,$y$ and $z$ coordinates, resulting in 3 additional feature maps. This representation aids the neural network in discerning the direction towards the goal and remains effective even when the goal lies outside the domain, as illustrated in the scenario depicted in  Fig.~\ref{fig:move_to_2_ball}. In total, there are seven input feature maps per controlled object.

\section{Training}\label{sec:training}

To train the system, we initialise objects with randomised locations and velocities, conduct simulations, and use the object trajectory for loss calculation. The loss function primarily expresses the distance of the object to the desired target location $\mathbf{x}^g$ with additional terms:
\begin{equation} \label{eq:loss}
    L = ||x_N - x^g||^2 +\alpha\sum_{k=1}^N ||x_k - x^g||^2+\beta\sum_{k=1}^N ||\mathbf{u}_k||^2.
\end{equation}
The first term ensures that the final object position is close to the target position. The second term keeps all intermediate positions near the target, effectively maximising the object's speed towards the target and maintaining its position once it arrives. We experimentally set $\alpha=1/N$ to achieve a good balance with the first and most important term. The third term aims to minimise the total energy expended by the emitters throughout the episode. We set a small weight $\beta=0.001$ for this term during training. The resulting reduction in fluid emissions based on this value is evaluated in Section~\ref{sec:results}.

All steps of the simulation are differentiable, and we employ gradient descent with the Adam optimiser \cite{kingma2015adam} using a learning rate of 0.0001. Based on experiments, the training is stable for values between 0.001 and 0.0001, resulting in minor differences in training speed. Training is performed for 2000 iterations, which takes about about 14 hours on a single Nvidia A100 graphics processor.

Training requires significant amount of GPU memory since all simulation steps must be preserved for backpropagation. With a $32\times32\times32$ domain discretization, approximately 80 steps can be stored on an A100 GPU. This is too small to complete the control tasks we consider. To enable longer rollouts, we employ truncated backpropagation through time, wherein backpropagation is applied solely to the final 80 steps of the rollout, while the total simulation length varies randomly from 80 to 280 steps. We observed that 280 training steps are sufficient to train the tasks, and the trained controller effectively supports much longer simulations, up to 600 steps used in evaluation, as shown in Section~\ref{sec:results}.

%% file: sections/3_results.tex
\section{Evaluation} \label{sec:results}
We evaluate the proposed system in three different scenarios that showcase the capabilities of our controller to carry out versatile object displacement using controlled fluid flows: 
\begin{itemize}
    \item[1.] keep the object in place, 
    \item[2.] move the object to a desired location, 
    \item[3.] independently control two objects. 
\end{itemize} 
Additionally, we investigate the system's robustness in response to external perturbations. During these evaluations, simulations are conducted over 600 time steps, which is sufficient to accomplish the specified tasks.

\begin{figure*}[htbp]
    \centering

    \hfill
    \subfigure[]{
        \includegraphics[width=0.47\textwidth]{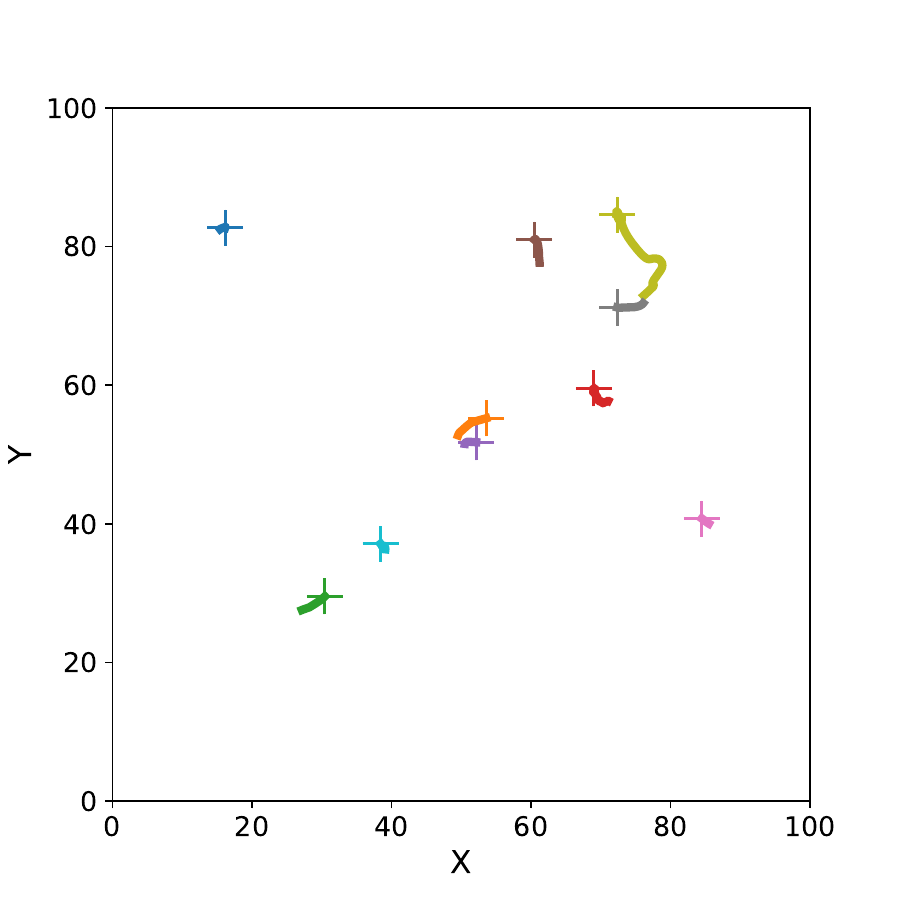} 
        \label{fig:keep_a}
    }
    \hfill
      \subfigure[]{
        \includegraphics[width=0.47\textwidth]{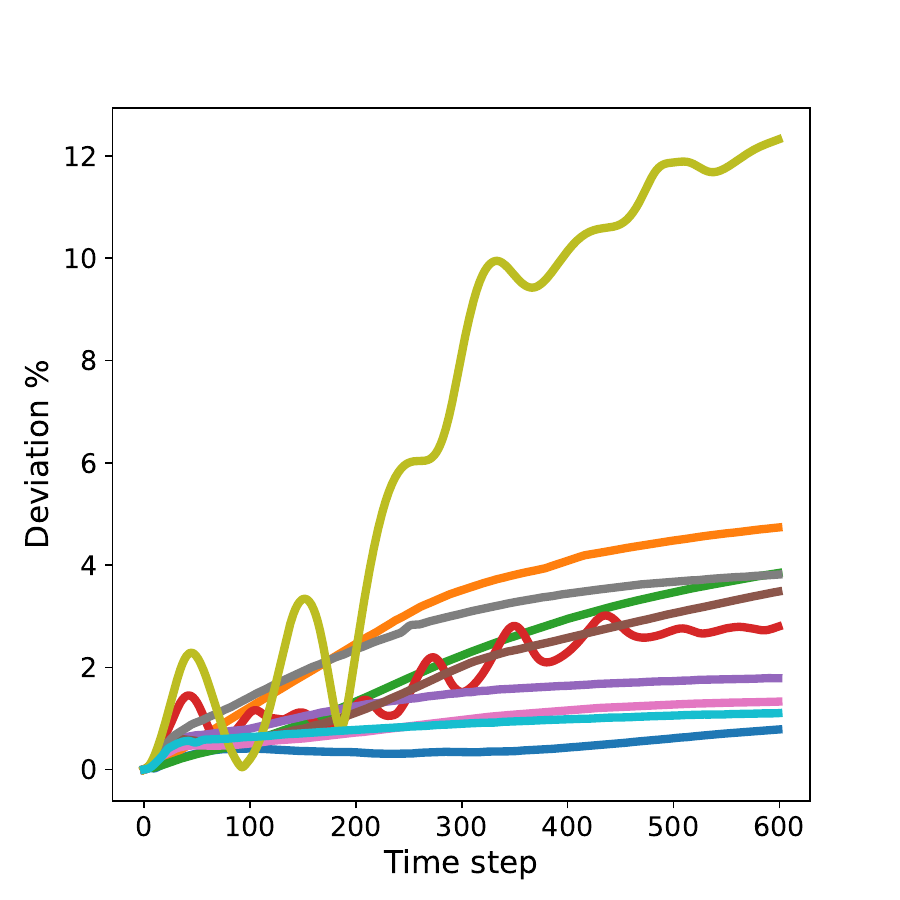} 
        \label{fig:keep_b}
    }

    \hfill
    \subfigure[]{
        \includegraphics[width=0.47\textwidth]{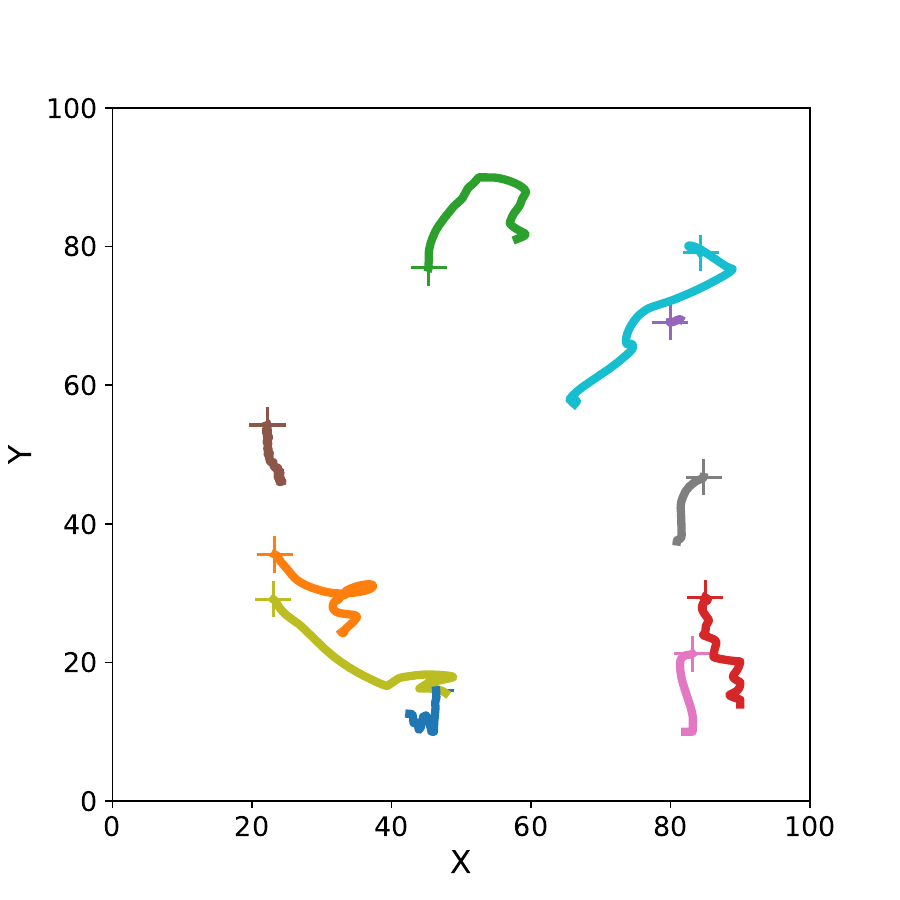} 
        \label{fig:wind_keep_a}
    }
    \hfill
    \subfigure[]{
        \includegraphics[width=0.47\textwidth]{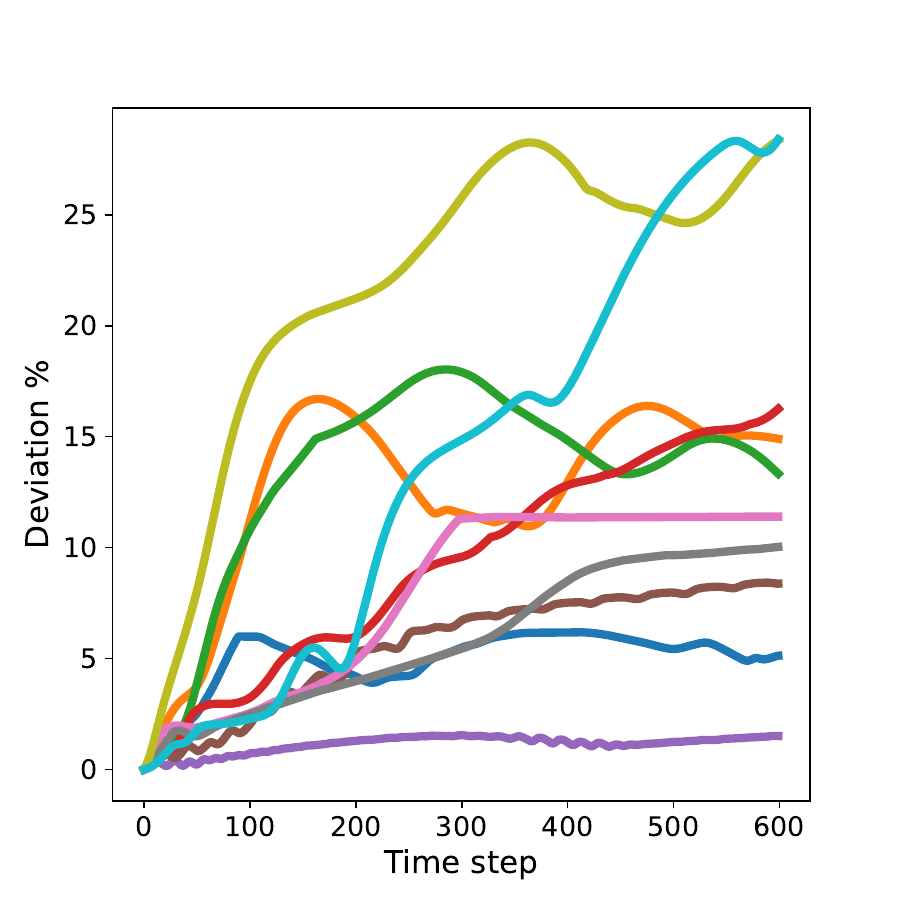} 
        \label{fig:wind_keep_b}
    }
    
    \hfill
     \caption{Task: maintain the position of the object without side wind (top) and with side wind (bottom). Left side: trajectories of several objects in the $xy$ plane. The target position is marked with a "+" sign. Right side: deviation from the target throughout the simulation time. The deviation is evaluated as the Euclidean distance to the desired location. Trajectory colours between left and right plots are set to match.}
    \label{fig:keep}
\end{figure*}

First, we consider the task of maintaining an object's position in a set location above the table. The object is initially placed at the desired location and the system has to learn to regulate the emitters to counteract the force of gravity, while also minimising horizontal displacement. Sample trajectories projected onto the $xy$ plane are depicted in Fig.~\ref{fig:keep_a}. In Fig.~\ref{fig:keep_b}, we measure the deviation from the target over time including the $z$ coordinate. The objects stay close to their initial position, within a tolerance of 5 units for all trajectories but one, where the units are relative the simulation domain where the table size is $100\times100$ units. However, we also observe that the error has a tendency to increase over time. This is caused by the fact that the controller is trained on 280 time steps and has to generalise for anything beyond that.

Next we analyse the system's robustness against external disturbances. To this end, we repeat the same scenario in the presence of side wind. A horizontal wind of random magnitude (up to 20\% of the emitter velocity) is applied during both training and testing. Its intensity is kept constant during a simulated episode. The controller does not know or observe the wind magnitude, so it must learn to counteract it solely from the observed ball motion. The results are depicted in Fig.~\ref{fig:wind_keep_a}. Again, the controller succeeds in learning this task, though at a reduced performance compared to the previous experiment. This is is an expected result, since the neural network has to acquire the wind strength by observing the ball motion before acting on it, and during this time the ball has already deviated slightly from its optimal location.

\begin{figure*}[htbp]
    \centering

    \subfigure[]{
        \includegraphics[width=0.30\textwidth]{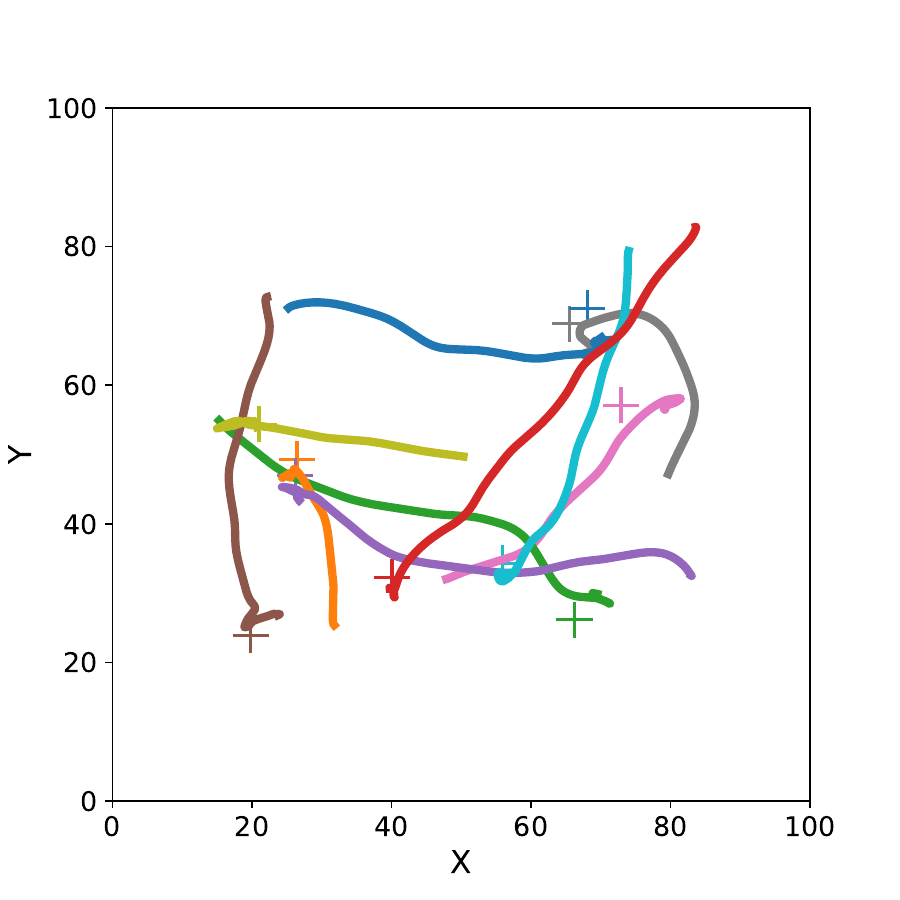} 
        \label{fig:move_aa}
    }
     \subfigure[]{
        \includegraphics[width=0.30\textwidth]{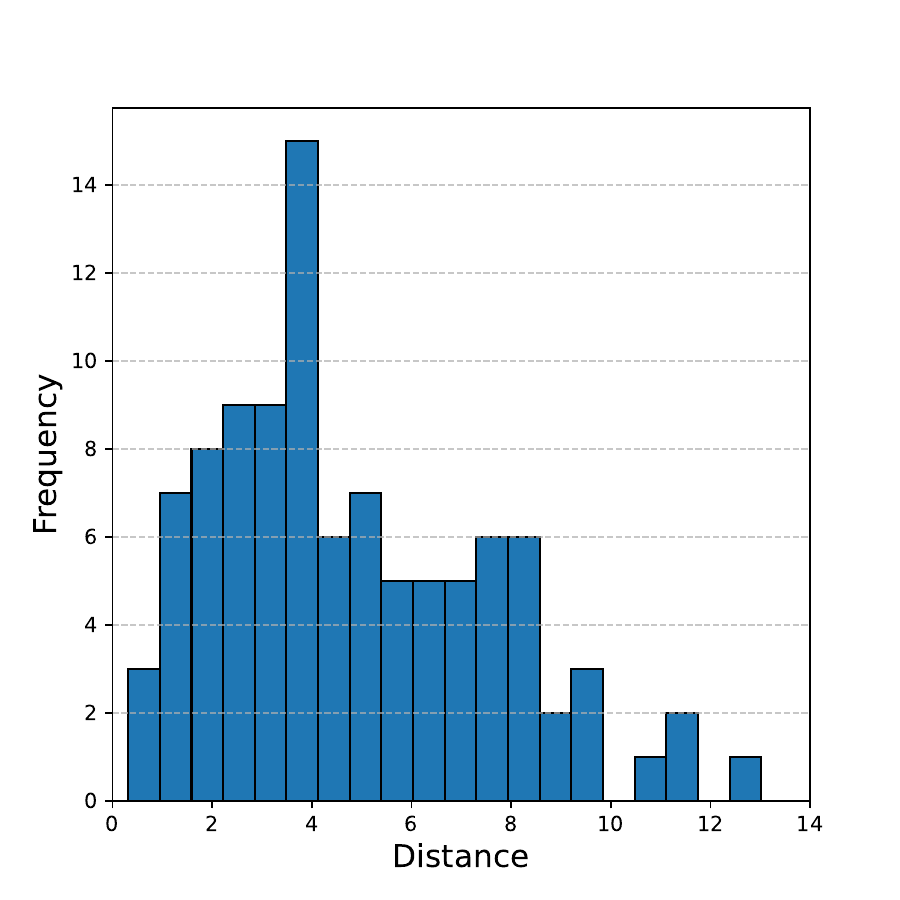} 
        \label{fig:move_bb}
    }
    \subfigure[]{
        \includegraphics[width=0.30\textwidth]{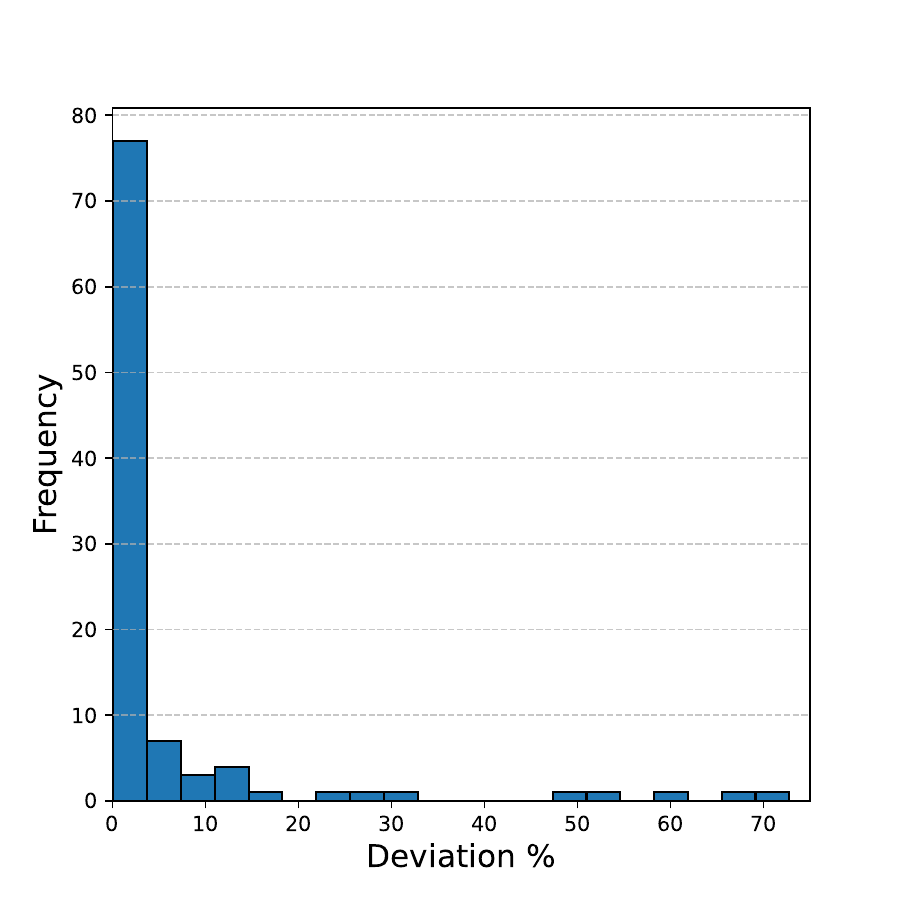} 
        \label{fig:move_cc}
    }
    \subfigure[]{
        \includegraphics[width=0.30\textwidth]{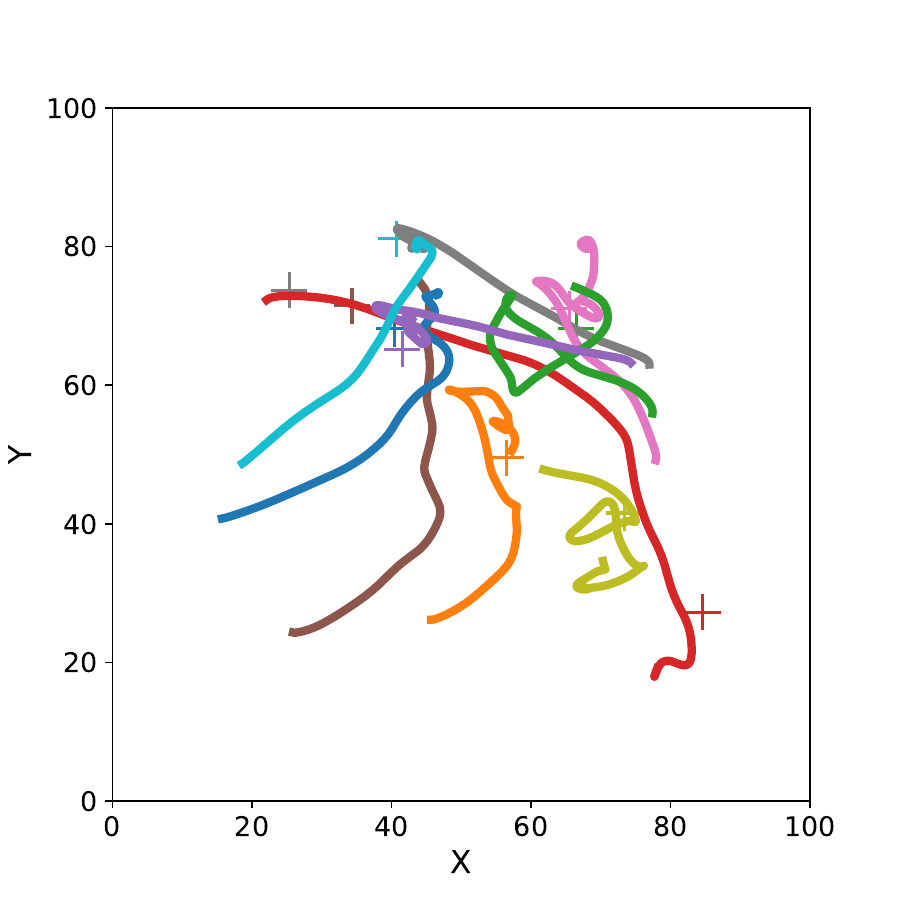} 
        \label{fig:move_wind_aa}
    }
     \subfigure[]{
        \includegraphics[width=0.30\textwidth]{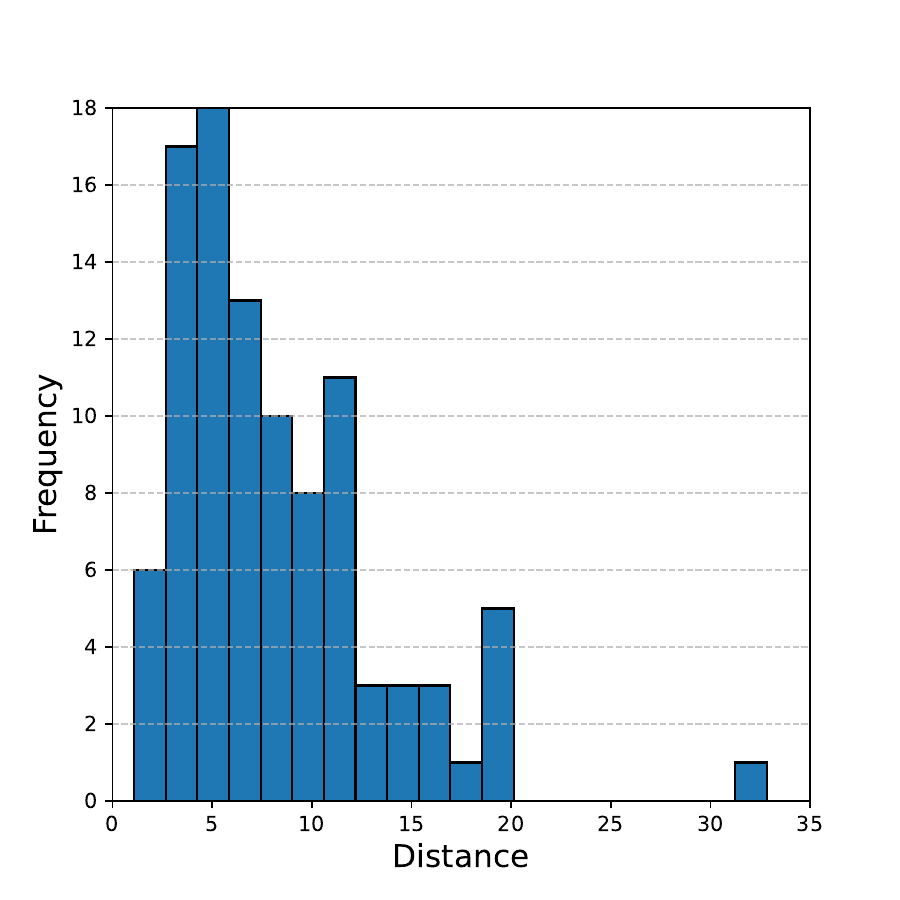} 
        \label{fig:move_wind_bb}
    }
    \subfigure[]{
        \includegraphics[width=0.30\textwidth]{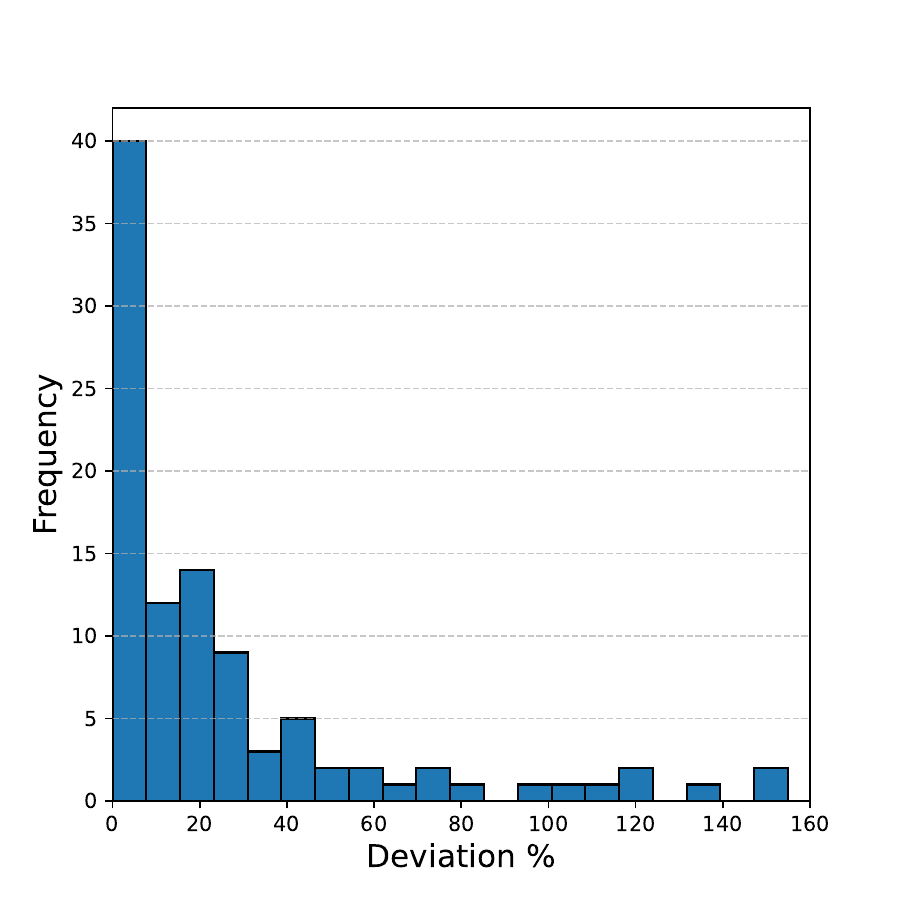} 
        \label{fig:move_wind_cc}
    }

    \hfill
    \caption{Task: move an object to the desired location. (a) Sample trajectories of the moved objects. The goal locations are marked with a "+" sign. (b) A histogram displaying proximity of the objects' location to the target position, measuring the distance between these two points in the last time step, for 100 trajectories. (c) Percentage by which the actual path deviates from a straight line, for 100 trajectories, shown as a histogram. (d) Sample trajectories of moved objects that are disturbed by side wind. (e) and (f) Trajectory accuracy metrics under the influence of side wind, analogous to (b) and (c).
    }
    \label{fig:move_100trajec}
\end{figure*}

The next task involves moving the object to a predetermined location. First, the object is randomly initialised on the table, and the task is to move it to another randomly specified location and hold it there until the end of the episode. A sample of the resulting trajectories is shown in Fig.~\ref{fig:move_aa}. In order to determine the accuracy of these trajectories, we measure the distance of the moved object to its target at the end of the episode. A histogram of results for 100 such measurements is shown in Fig.~\ref{fig:move_bb}. In most cases the object reaches a 5 unit distance from the target and the median distance is 4.0. 
Additionally, in Fig.~\ref{fig:move_cc}, we analyse the straightness of the trajectories, specifically comparing the length of the trajectory to the distance between the initial and final points. We see that the excess distance, shown as a percentage relative to the straight-line distance, is typically lower than 5\% with the median being 1.3.

We repeat the same experiment with side wind. The results are depicted in Fig.~\ref{fig:move_wind_aa}~-~\ref{fig:move_wind_cc}. We see that the controller is again able to guide the object towards the target, although with less precision (median distance from the target is 6.7 in 100 experiments) and with more curved trajectories (median straightness deviation is 14.4).

\begin{figure*}[htbp]
    \centering
    \hfill
    \subfigure[]{
        \includegraphics[width=0.47\textwidth]{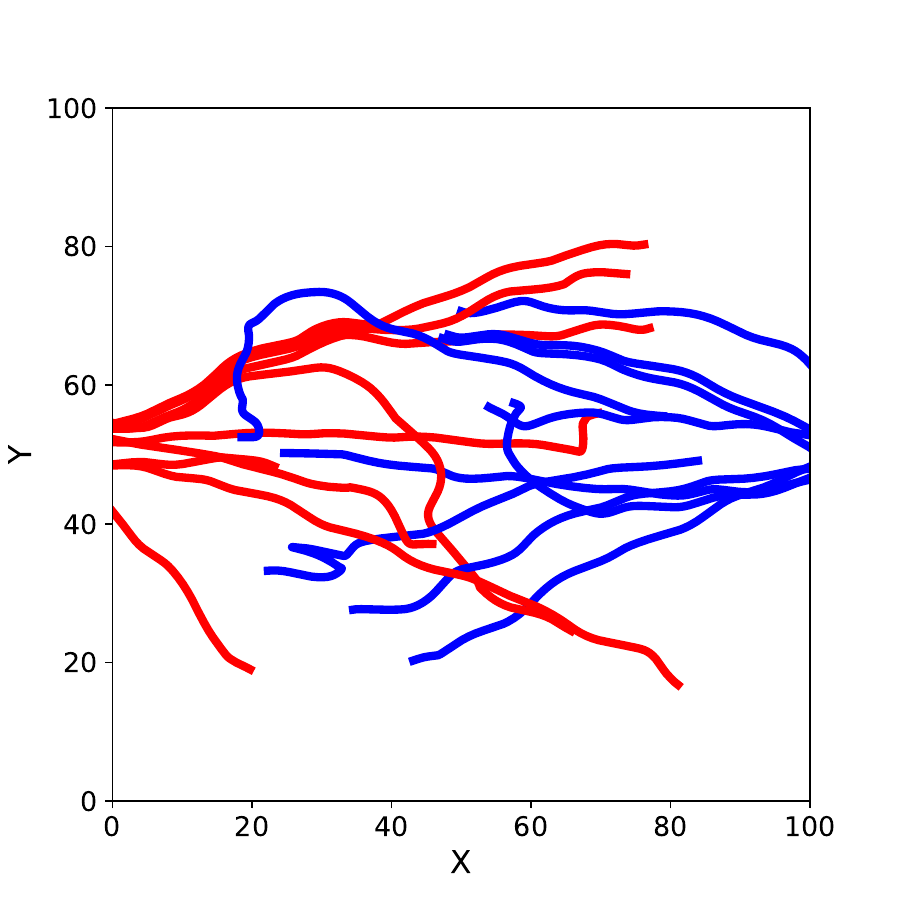} 
        \label{fig:move_to_2_ball_a}
    }
     \hfill
    \subfigure[]{
        \includegraphics[width=0.47\textwidth]{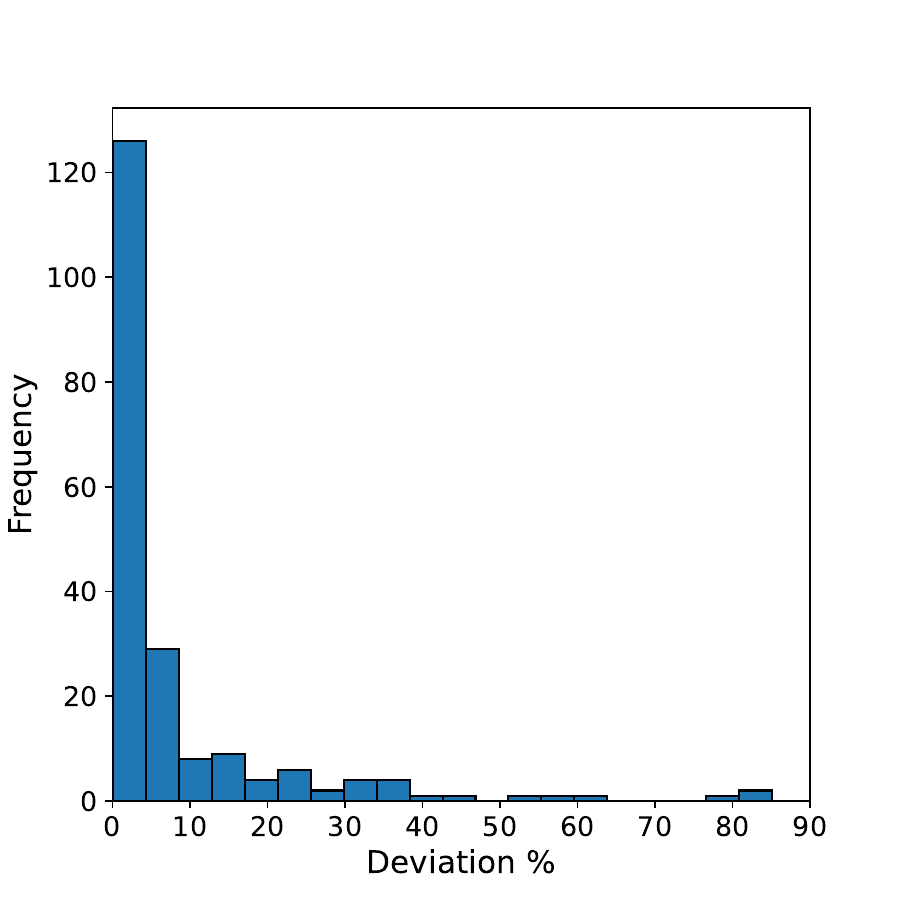} 
        \label{fig:move_to_2_ball_b}
    }
    \hfill
    \caption{Task: independently control two objects, where each object is required to arrive at a specified lateral position. (a) Sample trajectories of moved two objects, when the red object should reach the target point located at (-10, 50) on the left, and the blue object should reach the target point situated at (110, 50) on the right. (b) Percentage by which the actual path deviates from a straight line, for 100 trajectories.}
    \label{fig:move_to_2_ball}
\end{figure*}

Lastly, we investigate independent control of two objects. The task is to move the first object to the left of the table and the second one to the right simultaneously. The initial object locations are distributed randomly, meaning that their paths can collide. Results are shown in Fig.~\ref{fig:move_to_2_ball}. We see in Fig.~\ref{fig:move_to_2_ball_a} that the controller has learnt to correctly move both objects to their respective targets.
Additionally, when both objects come into close proximity, the emitters can separate them successfully and move them toward their respective targets, as seen in the \href{https://www.youtube.com/watch?v=sft0MH_pk9w}{video} provided in the supplementary material. In Fig.~\ref{fig:move_to_2_ball_b} we see the trajectory straightness depicted in the same way as in Fig.~\ref{fig:move_cc}. Compared with the single-object case, the straightness is slightly lower, mainly resulting from curved trajectories to avoid collisions.


In order to avoid high energy consumption for real-life implementations of such devices, it is generally advantageous to minimise the fluid consumption of the emitters. To that end, we increase the constant $\beta$ of the loss function in Eq.~\ref{eq:loss} to 0.02 and train the system for an additional 2000 iterations on top of the pre-trained model. Training it from scratch with the increased constant results in sub-optimal behaviour, with emitters turned off all the time. The chosen value of 0.02 is near the maximum threshold where the unwanted behaviour of emitters consistently shutting down, does not occur.

When tested on 100 simulations of the moving task, this adjustment results in an average reduction of emitted fluid volume from 0.32 to 0.05, while reducing the performance metrics only slightly. The average distance to the target increases from 3.32 to 4.74 but excess trajectory length increases from 5.28 to 7.78.

%% file: sections/4_conclusion.tex
\section{Conclusion} \label{sec:conclusion}

We have presented a method to train a neural network controller that can perform object displacement tasks using an array of fixed vertical fluid emitters. This approach has demonstrated versatile control tasks across multiple simulated scenarios, including: (1) hovering an object in place, (2) moving an object perpendicular to the flow direction, and (3) separating two objects. Unlike previous work, our approach relies solely on observing the object's state (its position and velocity) this way enabling to be implemented in real-world devices, where the object state can be readily monitored using computer vision techniques -- a direction we aim to explore further.

Our controller has also proven robust against external disturbances, which is an essential feature for bridging the gap between simulations and real-world devices. It also achieves object control while minimising the energy consumption in the emitters without significant loss in performance. The controller, trained by leveraging a differentiable simulation, reaches good performance within just 2000 iterations and can handle much longer episodes than those used during training.

However, several challenges remain before this approach can be implemented in a real-life device. The most important one is the expected difference in controller behaviour between simulation and reality. Currently, our simulations use abstract units for fluid and solid properties without corresponding to actual physical quantities. The simulation's resolution and episode lengths are also limited, which may not accurately reflect real-world dynamics. Moreover, we have only considered spherical objects without rotational inertia.

Despite these limitations, which are largely technical and addressable, our work represents a significant step toward developing real-world systems for manipulating objects with fluid streams.

%% file: bibliography.bib
@article{du2021_diffpd,
    author = {Du, Tao and Wu, Kui and Ma, Pingchuan and Wah, Sebastien and Spielberg, Andrew and Rus, Daniela and Matusik, Wojciech},
    title = {DiffPD: Differentiable Projective Dynamics},
    year = {2021},
    issue_date = {April 2022},
    publisher = {Association for Computing Machinery},
    address = {New York, NY, USA},
    volume = {41},
    number = {2},
    issn = {0730-0301},
    url = {https://doi.org/10.1145/3490168},
    doi = {10.1145/3490168},
    journal = {ACM Trans. Graph.},
    month = {11},
    articleno = {13},
    numpages = {21}
}

@inproceedings{holl2020learning,
  title={{Learning to Control PDEs with Differentiable Physics}},
  author={Holl, Philipp and Koltun, Vladlen and Thuerey, Nils},
  booktitle={Proceedings of the 8th International Conference on Learning Representations (ICLR)},
  year={2020}
}

@inproceedings{hu2019difftaichi,
  title={{DiffTaichi: Differentiable Programming for Physical Simulation}},
  author={Hu, Yuanming and Anderson, Luke and Li, Tzu-Mao and Sun, Qi and Carr, Nathan and Ragan-Kelley, Jonathan and Durand, Fr{\'e}do},
  booktitle={Proceedings of the 7th International Conference on Learning Representations (ICLR)},
  year={2019}
}

@article{ma2021diffaqua,
  title={Diffaqua: A differentiable computational design pipeline for soft underwater swimmers with shape interpolation},
  author={Ma, Pingchuan and Du, Tao and Zhang, John Z and Wu, Kui and Spielberg, Andrew and Katzschmann, Robert K and Matusik, Wojciech},
  journal={ACM Transactions on Graphics (TOG)},
  volume={40},
  number={4},
  pages={1--14},
  year={2021},
  publisher={ACM New York, NYamos2022control:, USA}
}

@article{fang2022complex,
  title={{Complex locomotion skill learning via differentiable physics}},
  author={Fang, Yu and Liu, Jiancheng and Zhang, Mingrui and Zhang, Jiasheng and Ma, Yidong and Li, Minchen and Hu, Yuanming and Jiang, Chenfanfu and Liu, Tiantian},
  journal={arXiv preprint arXiv:2206.02341},
  year={2022}
}

@inproceedings{teikmanis2023applying,
  title={{Applying a Differentiable Physics Simulation to Move Objects with Fluid Streams}},
  author={Teikmanis, Oskars and Leja, Laura and Freivalds, Karlis},
  booktitle={International Workshop on Embedded Digital Intelligence (IWoEDI)},
  year = {2023}
}

@inproceedings{ramos2022control,
  title={{Control of Two-way Coupled Fluid Systems with Differentiable Solvers}},
  author={Ramos, Brener and Trost, Felix and Thuerey, Nils},
  booktitle={ICLR 2022 Workshop on Generalizable Policy Learning in Physical World},
  year={2022}
}

@article{li2023difffr,
  title={DiffFR: Differentiable SPH-based Fluid-Rigid Coupling for Rigid Body Control},
  author={Li, Zhehao and Xu, Qingyu and Ye, Xiaohan and Ren, Bo and Liu, Ligang},
  journal={ACM Transactions on Graphics (TOG)},
  volume={42},
  number={6},
  pages={1--17},
  year={2023},
  publisher={ACM New York, NY, USA}
}

@article{tathawadekar2023incomplete,
  title={Incomplete to complete multiphysics forecasting: a hybrid approach for learning unknown phenomena},
  author={Tathawadekar, Nilam N and Doan, Nguyen Anh Khoa and Silva, Camilo F and Thuerey, Nils},
  journal={Data-Centric Engineering},
  volume={4},
  pages={e27},
  year={2023},
  publisher={Cambridge University Press}
}

@book{thuerey2021pbdl,
  title={Physics-based Deep Learning},
  author={Nils Thuerey and Philipp Holl and Maximilian Mueller and Patrick Schnell and Felix Trost and Kiwon Um},
  url={https://physicsbaseddeeplearning.org},
  year={2021},
  publisher={WWW}
}

@inproceedings{kingma2015adam,
  title={{Adam: A Method for Stochastic Optimization}},
  author={Kingma, Diederik P and Ba, Jimmy},
  booktitle={Proceedings of the 3rd International Conference on Learning Representations (ICLR)},
  year={2015}
}

@article{ma2018fluid,
  title={{Fluid directed rigid body control using deep reinforcement learning}},
  author={Ma, Pingchuan and Tian, Yunsheng and Pan, Zherong and Ren, Bo and Manocha, Dinesh},
  journal={ACM Transactions on Graphics (TOG)},
  volume={37},
  number={4},
  pages={1--11},
  year={2018},
  publisher={ACM New York, NY, USA}
}

@article{garnier2021review,
  title={A review on deep reinforcement learning for fluid mechanics},
  author={Garnier, Paul and Viquerat, Jonathan and Rabault, Jean and Larcher, Aur{\'e}lien and Kuhnle, Alexander and Hachem, Elie},
  journal={Computers \& Fluids},
  volume={225},
  pages={104973},
  year={2021},
  publisher={Elsevier}
}

@article{ren2022versatile,
  title={Versatile Control of Fluid-directed Solid Objects Using Multi-task Reinforcement Learning},
  author={Ren, Bo and Ye, Xiaohan and Pan, Zherong and Zhang, Taiyuan},
  journal={ACM Transactions on Graphics},
  volume={42},
  number={2},
  pages={1--14},
  year={2022},
  publisher={ACM New York, NY}
}

@article{fan2020reinforcement,
  title={Reinforcement learning for bluff body active flow control in experiments and simulations},
  author={Fan, Dixia and Yang, Liu and Wang, Zhicheng and Triantafyllou, Michael S and Karniadakis, George Em},
  journal={Proceedings of the National Academy of Sciences},
  volume={117},
  number={42},
  pages={26091--26098},
  year={2020},
  publisher={National Acad Sciences}
}

@article{verma2018efficient,
  title={Efficient collective swimming by harnessing vortices through deep reinforcement learning},
  author={Verma, Siddhartha and Novati, Guido and Koumoutsakos, Petros},
  journal={Proceedings of the National Academy of Sciences},
  volume={115},
  number={23},
  pages={5849--5854},
  year={2018},
  publisher={National Acad Sciences}
}

@article{rabault2019artificial,
  title={Artificial neural networks trained through deep reinforcement learning discover control strategies for active flow control},
  author={Rabault, Jean and Kuchta, Miroslav and Jensen, Atle and R{\'e}glade, Ulysse and Cerardi, Nicolas},
  journal={Journal of fluid mechanics},
  volume={865},
  pages={281--302},
  year={2019},
  publisher={Cambridge University Press}
}

@inproceedings{takahashi2021differentiable,
  title={Differentiable fluids with solid coupling for learning and control},
  author={Takahashi, Tetsuya and Liang, Junbang and Qiao, Yi-Ling and Lin, Ming C},
  booktitle={Proceedings of the AAAI conference on artificial intelligence},
  volume={35},
  pages={6138--6146},
  year={2021}
}

@inproceedings{he2016deep,
  title={Deep residual learning for image recognition},
  author={He, Kaiming and Zhang, Xiangyu and Ren, Shaoqing and Sun, Jian},
  booktitle={Proceedings of the IEEE conference on computer vision and pattern recognition},
  pages={770--778},
  year={2016}
}
